# Incorporating Discourse Aspects in English – Polish MT: Towards Robust Implementation

**Małgorzata E. Styś**
Computer Laboratory
University of Cambridge
New Museums Site
Pembroke Street
CB2 3QG Cambridge
England
mes1002@cl.cam.ac.uk

**Stefan S. Zemke**
Department of Computer
and Information Science
Linköping University
S-58183 Linköping
Sweden
steze@ida.liu.se

## Abstract

The main aim of translation is an accurate transfer of meaning so that the result is not only grammatically and lexically correct but also communicatively adequate. This paper stresses the need for discourse analysis in order to preserve the communicative meaning in English–Polish machine translation. Unlike English, which is a positional language with word order being grammatically determined, Polish displays a strong tendency to order constituents according to their degree of salience, so that the most informationally salient elements are placed towards the end of the clause regardless of their grammatical function.

The Centering Theory developed for tracking down given information units in English and the Theory of Functional Sentence Perspective predicting informativeness of subsequent constituents provide theoretical background for this work. The notion of *center* is extended to accommodate not only for pronominalisation and exact reiteration but also for definiteness and other center pointing constructs. Center information is additionally graded and applicable to all primary constituents in a given utterance. This information is used to order the post-transfer constituents correctly, relying on statistical regularities and some syntactic clues.

## 1 Introduction

Machine translation tends to concentrate on examining and conveying the meaning and structure of individual sentences. However, such action is not always sufficient. This paper discusses how analysis of intersentential connections could be performed and then exploited in MT. Such undertaking is thought to be necessary since the transfer of meaning has to be accurate not only on the lexical and grammatical level but also needs to carry across the communicative meaning of each utterance.

English and Polish exhibit certain idiosyncratic features which impose different ways of expressing the information status of constituents in succeeding clauses. Unlike English, which is a positional language with word order being grammatically determined, Polish displays a strong tendency to order constituents according to their degree of salience, so that the most informationally salient elements are placed towards the end of the clause regardless of their grammatical function. Such ordering of constituents yields solid knowledge about their degree of salience.

The paper is organised as follows. Section 2 includes a description of the center concept and an explanation of how it is carried across English clauses. A separate section is devoted to our extensions of the classic notion of center in view of machine translation. We then go on to describe the idiosyncratic properties of Polish and their implications for center transfer. Finally, practical rules for ordering Polish constituents are outlined.

The computational and theoretical background is supplied by the Centering Theory and the Theory of Functional Sentence Perspective. The former is used during English analysis while the latter provides theoretical framework for Polish gen-eration.

## 2 Centering Model for English Analysis

Centering as presented by Grosz, Joshi, Weinstein (Grosz et al. 86) and extended by Brennan, Friedman, Pollard (Brennan et al. 87) is a useful discourse model based on a system of rules for tracking down given information units on utterance level. Center, expressed as a noun phrase, is a pragmatic construct and it is intentionally defined as the discourse entity that the utterance is about.

### 2.1 Original Centering Algorithm

The current presentation of centering follows that by Grosz, et al. (Grosz et al. 86), (Grosz et al. 95). Identification of center is based on purely coreferential relations. Each utterance segment consists of utterances $U_1$, ..., $U_m$ and each of them exhibits one center. Associated with each utterance is a forward-looking center list $Cf(U_n)$ of all nominal expressions within $U_n$. The backward-looking center $Cb(U_n)$, which is the center proper, is the highest-ranked element of $Cf(U_n)$ realized in $U_{n-1}$. Pronominalisation and subjecthood are the main criteria underlying this ranking.

The entities on the Cf list are ordered by grammatical function which corresponds to the linear order of constituents in English. The first utterance in discourse has subject as its center by default. Generally, however, resolvable pronouns are the preferred center candidates, since they are the most common devices signalling the relation of coreference. Comparison of centers can generally relate utterances in the way of continuing ($Cb(U_n) = Cb(U_{n-1})$) or shifting ($Cb(U_n) \neq Cb(U_{n-1})$). For definition of more subtle relations look at (Brennan et al. 87).

### 2.2 Extension of the Algorithm

Various refinements have been added to the original centering model since its introduction (Brennan et al. 87), (Kameyama 86), (Mitkov 94), (Walker et al. 94). Center identification is mostly based on syntactic phenomena. It concentrates on the analysis and representation of noun phrases, since the function of nominals in the information structure is considered to be crucial.

Below we present some of our extensions to centering,

- Additional criteria for center evaluation
  - Special center-pointing constructions
  - Demonstrative pronouns
  - Possessive and demonstrative modifiers
  - Extra credit for definite articles
  - Indefinites decreasing center value of a constituent

- Gradation of center values

- Center values given to all Nps (not just one)

- Composite computation of a center value depending on a number of clues

- Introduction of a limited referential distance which depends on constituent length

- Incorporation of synonyms in reiteration detection

We choose the constituent with the highest center value as the discrete center of an utterance. If more than one constituent has been assigned the same value, we take the entity that is highest-ranked according to the ranking introduced by (Grosz et al. 86), (Grosz et al. 95), (Brennan et al. 87).

Within the next few sections, we provide a description of those centering criteria that have been added to the original algorithm.

#### 2.2.1 Definiteness

Definite noun phrases are often co-specifiers of current centers. The correlation between definiteness and an entity having been introduced in previous discourse in English is high but not total. (For example, proper names can be textually new yet definite.) We therefore include definiteness among factors contributing to center evaluation. Indefinite noun phrases are treated as new discourse entities.

#### 2.2.2 Lexical Reiteration

Lexically reiterated items include repeated or synonymous noun phrases often preceded by definite articles, possessives or demonstratives. We also propose to consider semantic equivalence based on the synonyms coded in the lexicon as valid instances of reiteration.

#### 2.2.3 Referential Distance

For pronouns and reiterated nouns, we propose the allowed maximal referential distance, measured in the number of clauses scanned back, to correlate with the word length of the constituent involved (Siewierska 93b). This relates to the observation that short referring expressions have their resolvents closer than longer ones. Such precaution limiting the referential distance minimises the danger of over-interpretation of common generic expression such as *it*.

Although we haven't yet experimented with various functions relating the type of referent to its allowed referential distance, a simple linear dependence (with factor 1-2) seems to be reasonable. Thus, in the following example, we will assume the referential distance to be twice the length of the (resulting Polish) constituent.

### 2.2.4 Center-pointing Constructions

Certain English constructions unambiguously point to the center thus making more detailed analysis unnecessary.

Although the subject is obligatory in an English sentence, occasionally a formal slot-filling item is substituted in its place giving rise to a cleft construction. The information structure becomes explicit by virtue of the fact that it exhibits a structurally marked center. (Eg. It was John who came.) The center could also be fronted (Eg. Apples, Adam likes) or introduced using a sort of sentence equivalent (Eg. As for Adam, he doesn't like apples). Sidetracking from the main thread of discourse is a common device used by the speaker to direct the attention of the addressee. Expressions such as *as for*, *concerning*, *with regard to* are such prompts.

### 2.2.5 Center Gradation

Considering the priority scale of referential items, the mechanisms underlying centering in English could then be outlined as follows,

- Preference of pronouns over full nouns
- Preference of definites over indefinites
- Preference of reiterated items over non-reiterated ones
- Preference of constituents involving more "givenness" indicators

These considered along with special center-pointing constructions give rise to the following numerical guidelines (some of which agree with the idea of a givenness hierarchy cf. (Gundel 93)),

1. Highest center value is given to "unquestionable" centers:
   - pointed out by center-indicating constructions such as clefts,
   - according to (Grosz *et al.* 86): (resolved) personal pronouns,
   - consisting of (definite) reiterated possessive expressions.

2. Lower priority is given to
   - definite reiterations,
   - other resolved pronouns,
   - not fully reiterated or definite possessive expressions,
   - demonstratives.

3. The lowest positive value is given to dubious centers such as
   - non-definite reiterations,
   - non-iterated definites.

4. Value -1 given to new information units introduced by
   - indefinite articles "a/an",
   - determiners "another", "other".

5. Neutral value 0 is assigned to all other Np.

The rules for Composite Centers allow us to calculate center value increase over the default value 0. Thus, for example, the center value for *the scientists' colleagues* will be arrived at by adding the contribution for *the* (+1) to the contributions for *scientists* and *colleagues* (each 0 or 1 depending on whether the item is reiterated) giving a value between 1 and 3 depending on the context. In Figure 2, we illustrate the application of rules included in Figure 1.

The assumtion for all center rules is that the highest possible center value is derived.

## 3 Local Discourse Mechanisms in Translation

In discourse analysis, we relate particular utterances to their linguistic and non-linguistic environment. Below, we shall describe the relationship between the grammatical sentence pattern (Subject Verb Object) and the communicative pattern (Theme Transition Rheme).

### 3.1 Functional Sentence Perspective

FSP is an approach used by the Prague School of linguists to analyse utterances of Slavic languages in terms of their information content (Firbas 92). In a coherent text, the given or known information, *theme*, usually appears first thus forming a co-referential link with the preceding text. The new information, *rheme*, provides some information about the theme. It is the essential piece of information of the utterance.

There are clear linear effects of FSP[1]. Utterance non-final positions usually have given information interpretation and the final section of the utterance represents the new. This phenomenon could be explained by word order arranged in such a way that first come words pointing to details already familiar from the preceding utterances/external context and only then come words describing new detail. Similarly, in the process of mental activities first comes the process of identification and then augmentation of received perception. It is then followed by details individually connected with the given idea (Szwedek 76).

---

[1] The information structure also changes depending on the accentuation pattern, but we shall leave the intonation aspects aside in this presentation.

|   | SELECTION CRITERIA | SYNTACTIC MARKERS | CENTER VALUE |
|---|---|---|---|
|   | Center-pointing constructions (Point.1-3) | | |
| 1 | Cleft | it+Be+$N_c$+that/who | center($N_c$):=3 |
| 2 | Fronted | $N_f$,Sentence-$N_f$ | center($N_f$):=3 |
| 3 | Prompted | Prompt+$N_p$,Sentence | center($N_p$):=3 |
|   | Pronominal centers (Pron.1-3) | | |
| 1 | Personal (resolved) | I/you/it/he/she/we/they | center($Pron_{pers}$):=3 |
| 2 | Demonstrative (resolved) | this/that/these/those | center($Pron_{demo}$):=2 |
|   | Non-centers (Non.1-2) | | |
| 1 | Indefinites | a/an/another/other | center($N_{indef}$):=-1 |
| 2 | Default for any Np | *Cases not listed elsewhere* | center(Np):=0 |
|   | Composite Centers (Comp.1-4) | | CENTER INCREASE |
| 1 | Reiterated nominals | $N_{reit} \overset{ref\_dist}{\longleftrightarrow} N_{reit}$ | center($N_{reit}$)+1 |
| 2 | Definite expressions | the/such/this/that/these/those +N | center(N)+1 |
| 3 | Possessives | its/his/her/our/your/their +N | center(N)+2 |
| 4 | Genitives | $N_o$'s+$N_p$, $N_p$+of+$N_o$ | center($N_p$)+center($N_o$) |

Figure 1: Center values for different types of NP

| No. | UTTERANCE | RULES | VALUES | CENTER |
|---|---|---|---|---|
| 1 | The scientists conducted many tests. | Comp.2 Non.2 | 1 = 1+0 0 | scientists |
| 2 | The tests were thorough. | Comp.1,2 | 2 = 1+1+0 | tests |
| 3 | The results were examined by their colleagues. | Comp.2 Comp.3 | 1 = 1+0 2 = 2+0 | colleagues |
| 4 | They were judged convincing. | Pron.1 | 3 | they = results |
| 5 | The scientists' colleagues were impressed by the tests. | Comp.1,2,4 Comp.1,2 | 3 = 1+1+1+0+0 2 = 1+1+0 | colleagues |

Figure 2: Center values for example clauses

### 3.2 Constituent Order in Polish Translation

The distribution of new/old information determines the order of constituents within clauses. Since the grammatical function is determined by inflection in Polish, there is great scope for constituent order to express contextual distinctions and the order often seems free due to virtual absence of structural obstacles.

Just as it is not valid to assign any order to a sequence of constituents, one cannot keep repeating the SVO sequence in all cases. If we were to translate all sentences of an English text into Polish following the canonical SVO pattern, we would get a grammatically correct but often communicatively inadequate and incoherent text. Thus in order to decide which ordering to use, we have to take into consideration the commonly occurring contextual functions and their implications for the probability and the frequency of occurrence of a given order. The degree of emphasis is also a factor and it is worth noting that the more frequently an order occurs the less emphatic it is (Siewierska 93b).

Restrictions on phrasal constituent order can be broadly placed under three categories: contextual, grammatical and stylistic. The grammatical restrictions are not as strict as in English and the stylistic constraints are omitted within the scope of this paper. The remaining sections concentrate on the former two categories.

## 4 Ordering of Polish Constituents

### 4.1 The Ordering Approach

As it has already been argued, center information is crucial for the communicatively correct positioning of Polish constituents in a flow of text. However, there are other factors influencing the order which can co-specify or even override it. This presents a delicate task of balancing a number of clues selecting the most justified order(s) or – in the case of a discriminating approach – the ones which do not have any strong arguments against them.

Our choice of ordering criteria has been directly based on the findings of the Prague School discussed above, our own linguistic experience (both of us bilingual, native speakers of Polish) and on some statistical data provided by (Siewierska 93b), (Siewierska 93a), (Siewierska 87).

The intended approach to ordering could be characterised as follows,

**Permissive:** Generate more (imperfect) versions rather than none at all. If need be, restrict by further filters.

**Composite:** Generate all plausible orders before some of them will be discriminated. (This approach is side-tracked when a special construction is encountered.)

**Discrete:** No gradings/probability measures are assigned to competing orders as to discriminate between them. This could be an extension.

### 4.2 Ordering Criteria

The ordering of constituents in Polish utterances generally follows the communicative order from given to new. Below we present some rules which are obeyed by Polish clauses under normal conditions,

- End weight principle: Last primary constituent is the anti-center.
- Given information fronting: Constituents belonging to the given information sequence are fronted.
- Short precede long principle: Shorter constituents go first.
- Relative order principle: Certain partial orders are only compatible with specific patterns of constituents.

Additionally, there is a strong tendency to omit subject pronouns. Such omission, however, exhibits different degrees of optionality.

What follows is a list of constructs used in subsequent tables to generate plausible orders of (translated) Polish constituents.

**Center information:** has the highest rank in the ordering procedure and is used in two aspects:
- center(*Constituent*) returns the center value of the *Constituent*'s Np, or 0 if undefined,
- center_shift(*Utterance*) holds if *Utterance* relates to the preceding one in the way allowed by the shift transition cf. (Grosz et al. 86)
- discrete_center(*Constituent*) holds if *Constituent* is the chosen center of the current utterance

**Length of constituents:** length(*Constituent*) returns the number of words of the resulting Polish *Constituent*[2]. Although not as important as center information, this rough measure can discriminate certain orders on the basis of "short precedes long" principle[3].

---

[2] To a great extent, this measure depends on the translation of constituents. It could be simplified by measuring the length of the original English, instead of Polish, units. We make use of that simplified measure in the example

**Positioning of certain constituents:** (or indeed their lack) can in turn induce other constituents to occupy certain positions. Some orders are only possible in certain configurations, e.g. with frontal *Adjunct* (X-), whereas others require just its presence (-X-), or absence (X=[ ]).

**Syntactic phenomena:**

- grammatical function of a constituent, eg. being a subject (S) or object (O)
- pron(*S*) & pron(*O*) if both subject and object are pronominal or Sub($U_n$) = Sub($U_{n-1}$) – if subject stays the same.
- certain expressions, e.g. a focus binding expression such as 'only', can trigger specific translation patterns.

**Features of next utterance:** e.g. center(S,$U_{n+1}$) > 0, can be used together with the features of the current utterance in order to obtain more specific conditions.

In the following tables S denotes (Polish) subject, V – verb, O – object, X – adjunct, Prim – S or O, "-" – (sequence of) any, [ ] – omitted constituent. The difference for "≫" to hold must be at least 2.

### 4.3 Building on Orders of Constituents

The Preference Table presents some of the main PREFERENCEs for generating orders of Polish constituents depending on CONDITIONS. Each line of the table can be treated as an independent if-then rule co-specifying (certain aspects of) an order. CONDITIONS being simple conjunctions (of regular expressions) are intended to allow straightforward transformation into a Prolog program. Different rules can be applied independently thus possibly better determining a given order[4]. The JUSTIFICATION column provides some explanation of the validity of each rule; 'bare' indicates the percentage of bare constructions including three primary constituents only. Both the Preference Table and the Discrimination Table are mostly based on statistical data gathered by (Siewierska 87), (Siewierska 93b), (Siewierska 93a).

### 4.4 Discriminating Orders

It might be the case that as a result of applying the Preference Table, we obtain too many orders. The Discrimination Table provides some rationale for excluding those matching ORDERs

---
[3]It is interesting to note, however, that for the otherwise rare order OSV, the opposite applies.

[4]Orders derived by co-operation of several rules could be preferred in some way.

---

for which one of their DISCRIMINATION conditions fails. If the building stage left us with no possible orders at all, we could allow any order and pick only those which successfully pass all their discrimination tests. It is purposeful that all orders apart from the canonical SVO have some discrimination conditions attached to them. The rarer the order tends to be the more strict the condition. Therefore, SVO is expected to be the prevailing order.

### 4.5 Special Cases

There remains a number of cases which escape simple characterisation in terms of "preferred and not-discriminated". The Preprocessing Table offers some solutions under such circumstances. It is to be checked for its conditions before any of the previous tables are involved. If a condition holds, its result (eg. 0-anaphora) should be noted and only then the other tables applied to co-specify features of the translation as described above. The Preprocessing Table can yield erroneous results when applied repeatedly for the same clause. Therefore, unlike the other tables, it should be used only once per utterance.

### 4.6 Example

In Figure 6 we continue the example from Figure 2. The orderings built on by a cooperation of the Preprocessing/Preference and not refused by the Discrimination Table appear in the last column.

## 5 Conclusion

One of the aims of this research was to exploit the notion of center in Polish and put it forward in context of machine translation. Centers are conceptualised and coded differently in Polish and English utterances. This fact has clear repercussions in the process of translation. Through exploring the pragmatic, semantic and syntactic conditions underlying the organisation of utterances in both languages, we have been able to devise a set of rules for communicatively motivated ordering of Polish constituents.

Among the main factors determining this positioning are pronominalisation, lexical reiteration, definiteness, grammatical function and special centered constructions in the source language. Their degree of topicality is coded by the derived center values. Those along with additional factors, such as the length of the originating Polish constituents and the presence of adjuncts, are used to determine justifiable constituent order in the resulting Polish clauses.

In future research, we wish to extend the scope of translated constructions to di-transitives and passives. We shall also give due attention to relative clauses. Centering in English can be further

| Pref. | CONDITIONS | PREFERENCE | JUSTIFICATION |
|---|---|---|---|
| \multicolumn{4}{c}{Orderings implied by center information} | | | |
| i | center(Any) < 0 | -Any | Final position of new |
| ii | center(Any1) ≫ center(Any2) | -Any1-Any2- | Given-new principle |
| iii | center(X) > 1 | X- | Adjunct topic fronted |
| iiib | discrete_center(Prim) | (X-)(V-)Prim- | Primary center fronted |
| \multicolumn{4}{c}{Statistical positioning preferences} | | | |
| iv | -V-S-O- & -X- | XV-S-O- | Statistical (66%; bare 11%) |
| v | -O-S- & X- | XV-O-S- | Statistical |
| vi | -V-O-S- & -X- | XV-O-S- | Statistical (53%; bare 28%) |
| vii | -S-V-O- & -X- | XS-V-O- | Statistical (32%; bare 18%) |
| viii | -S-V-O- & -X- | S-V-OX | Statistical (30%; bare 18%) |
| ix | -O-V-S- & -X- | O-V-SX | Statistical (29%; bare 28%) |
| x | -O-V-S- & -X- | O-VXS | Statistical (26%; bare 28%) |
| xi | pron(S) (& center_shift($U_n$) ) | -VS- | Stylistic |
| xii | *No condition so preference* | -V-O- | Statistical (89%+) |
| xiii | *weaker than any other* | -S-O- | Statistical (81%) |

Figure 3: Preference Table

| Discr. | ORDER | DISCRIMINATION | JUSTIFICATION |
|---|---|---|---|
| i | -V-S-O- | length(S) ≤ length(O) | Statistical (99%) |
| ii | -V-S-O- | -V-S-O | Statistical (87%) |
| iii | -V-S-O- | Pron(S) | Stylistic |
| iv | -V-O-S- | length(O) ≤ length(S) | Statistical (96%) |
| v | -V-O-S- | -X- present | Statistical (89%) |
| vi | -S-O-V- | SOV | Statistical (50%+) |
| vii | -S-O-V- | center(S,$U_{n+1}$) > 0 | Statistical |
| viii | -O-S-V- | OSVX | Statistical (79%) |
| ix | -O-S-V- | length(O) ≥ length(S) | Statistical (100%) |
| x | -O-V-S | length(O) ≥ length(S) | Statistical (64%) |

Figure 4: Discrimination Table

| Pre. | CONDITIONS | RESULT | JUSTIFICATION |
|---|---|---|---|
| \multicolumn{4}{c}{0-anaphora} | | | |
| i | S='we' | S=[ ] | Rhythmic |
| ii | pron(O) & pron(S) | S=[ ] | Stylistic |
| iii | Sub($U_n$) = Sub($U_{n-1}$) (& pron(S)) | S=[ ] | Stylistic |
| iv | center_continuing($U_n$) | S=[ ] | Stylistic |
| \multicolumn{4}{c}{Special constructions} | | | |
| v | -'only' SV- & pron(S) | -'tylko' SV- | Focus binding expr. |
| vi | X=[ ] & pron(O) | SOV | Special: S,O,V only |

Figure 5: Preprocessing Table

| Utter-ance | PREFFERENCE CRITERIA | PARTIAL ORDERINGS | DISCRIMINATION (FAILING) | RESULTING ORDER(S) |
|---|---|---|---|---|
| 1 | Pref.xii<br>Pref.xiii | SVO<br>VSO | (Discr.iii) | SVO |
| 2 | *No rules apply, order unchanged* | | | SVX |
| 3 | Pref.iiib<br>(Pref.xii) | OVS<br>VOS<br>OSV | Discr.x<br>(Discr.v)<br>(Discr.viii) | OVS |
| 4 | Pre.iii<br>Pref.xi | S=[ ]<br>-VS- | | V[S]X |
| 5 | Pref.iiib<br>(Pref.xii) | SVO<br>VSO | (Discr.i) | SVO |

Figure 6: Example continued: Deriving constituent orders

refined by allowing verbal and adjectival centers as well as by determining anti-center constructs.

We have thus tackled the question of information distribution in terms of communicative functions and examined its influence on the syntactic structure of the source and target utterances. How and why intersentential relations are to be transmitted across the two languages remains an intricate question, but we believe to have partially contributed to the solution of this problem.